\documentclass{article}

\PassOptionsToPackage{numbers, compress}{natbib}

\usepackage[preprint]{neurips_2026}

\usepackage[utf8]{inputenc}
\usepackage[T1]{fontenc}
\usepackage{hyperref}
\usepackage{wrapfig}
\usepackage{subcaption}
\usepackage{url}
\usepackage{booktabs}
\usepackage{amsmath}
\usepackage{amsfonts}
\usepackage{nicefrac}
\usepackage{microtype}
\usepackage{float}
\usepackage{CJKutf8}
\usepackage[table,dvipsnames]{xcolor}
\usepackage{xspace}
\usepackage{marvosym}
\usepackage{multirow}
\usepackage{enumitem}
\usepackage{makecell}
\usepackage{pifont}
\usepackage{hhline}
\usepackage{nicematrix}
\usepackage{tikz} 
\usepackage{makecell}
\definecolor{TableHeaderBlue}{RGB}{193,216,224}
\definecolor{TableYesGreen}{RGB}{46,125,50}
\definecolor{TableNoRed}{RGB}{198,40,40}

\usepackage{graphicx}
\newcommand{\huggingface}{\includegraphics[height=1em]{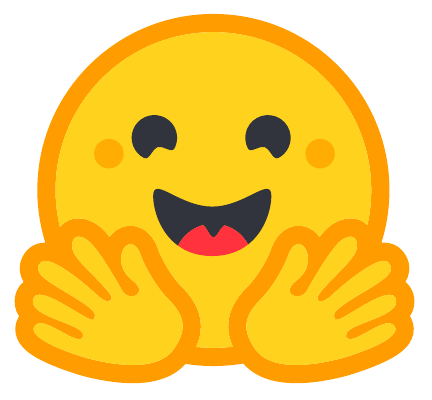}\xspace}
\newcommand{\github}{\includegraphics[height=1em]{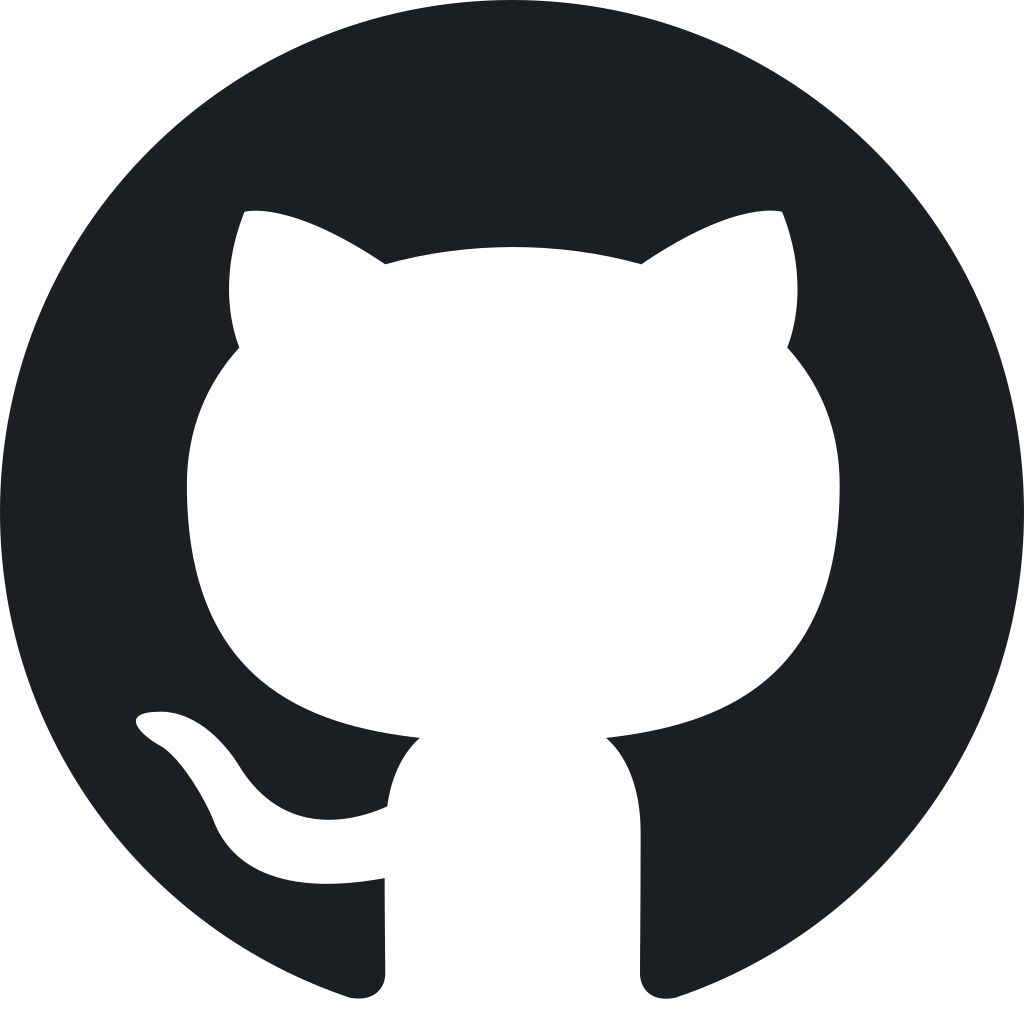}\xspace}

\hypersetup{
    colorlinks = true,
    linkbordercolor = {white},
    linkcolor = blue!60!black,
    citecolor = blue!60!black,
    urlcolor = blue!60!black
}

\title{\hspace{-0.4em}
\raisebox{-0.3\height}{\includegraphics[height=1cm]{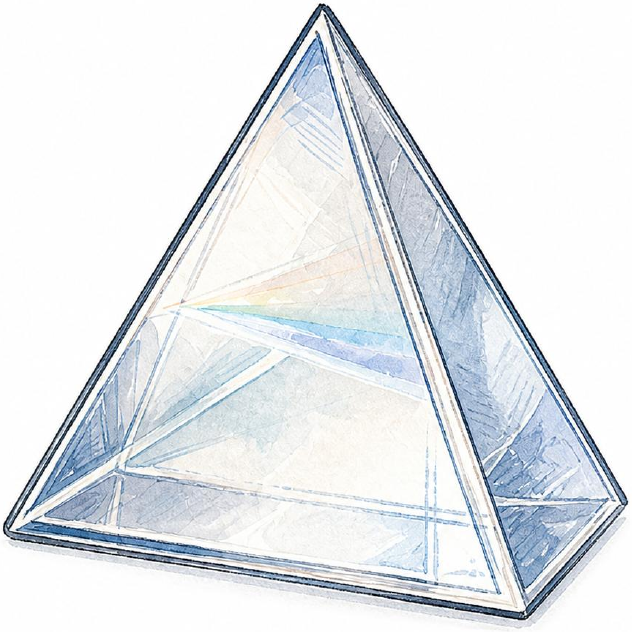}} 
PRISM: A Benchmark for Programmatic Spatial-Temporal Reasoning
}

\author{%
\textbf{
\begin{tabular}{c}
Qiran Zhang$^\dagger$ \quad Yuheng Wang$^\dagger$ \quad Runde Yang \quad Lin Wu \quad
Jingru Fan \quad Shu Yao \\ Jie Zhang \quad Tianle Zhou \quad Huatao Li \quad Ruijie Shi \quad Yihan Li \quad Chen Qian\textsuperscript{\Letter}
\end{tabular}
} \\
  \vspace{-0.5em} \\
  \textnormal{School of Artificial Intelligence, Shanghai Jiao Tong University}
  \vspace{0.4em} \\
    \texttt{\{qrzhang\_23,qianc\}@sjtu.edu.cn}
  \vspace{0.4em} \\
  \github \textbf{Code}:~\texttt{\href{https://github.com/positionprivacy/PRISM}{github.com/positionprivacy/PRISM}}
  \vspace{0.1em} \\
  \huggingface \textbf{Dataset}:~\texttt{\href{https://huggingface.co/datasets/posprivacy/PRISM}{datasets/posprivacy/PRISM}}
}

\begin{document}

\maketitle
\begingroup
\renewcommand\thefootnote{}
\footnotetext{$^\dagger$ Equal contribution.}
\footnotetext{\Letter\ Corresponding author.}
\endgroup
\setcounter{footnote}{0}

\vspace{-2em}
\begin{abstract}
Programmatic video generation through code offers geometric precision and temporal coherence beyond pixel-level diffusion models, yet rigorously evaluating whether language models can produce spatially correct animated outputs remains an open problem. We introduce \textbf{PRISM}, a large-scale benchmark of 10,372 human-calibrated instruction-code pairs ($20\times$ larger than prior programmatic video generation benchmarks), grounded in real-world knowledge visualization scenarios across English and Chinese and spanning 437 subject categories. We further propose a funnel-style evaluation framework with four complementary metrics: \emph{Code-Level Reliability} for executability, \emph{Spatial Reasoning} for layout correctness over full animation sequences, and \emph{Prompt-Aware Dynamic Visual Complexity} (PADVC) and \emph{Temporal Density} (TD) for diagnosing dynamic expression and temporal activity. Systematic evaluation of seven mainstream LLMs reveals a striking \emph{Execution-Spatial Gap}: the average drop from execution success rate to spatial pass rate is approximately 41\%, showing that runnable code does not necessarily yield spatially coherent visual output. These findings show that programmatic video generation evaluation should go beyond executability. PRISM provides a principled benchmark for advancing spatially coherent code generation.
\end{abstract}

\section{Introduction}

\begin{wrapfigure}{r}{0.53\columnwidth}
    \vspace{-1.0em}
    \centering

    \begin{subfigure}[t]{0.48\linewidth}
        \centering
        \includegraphics[width=\linewidth]{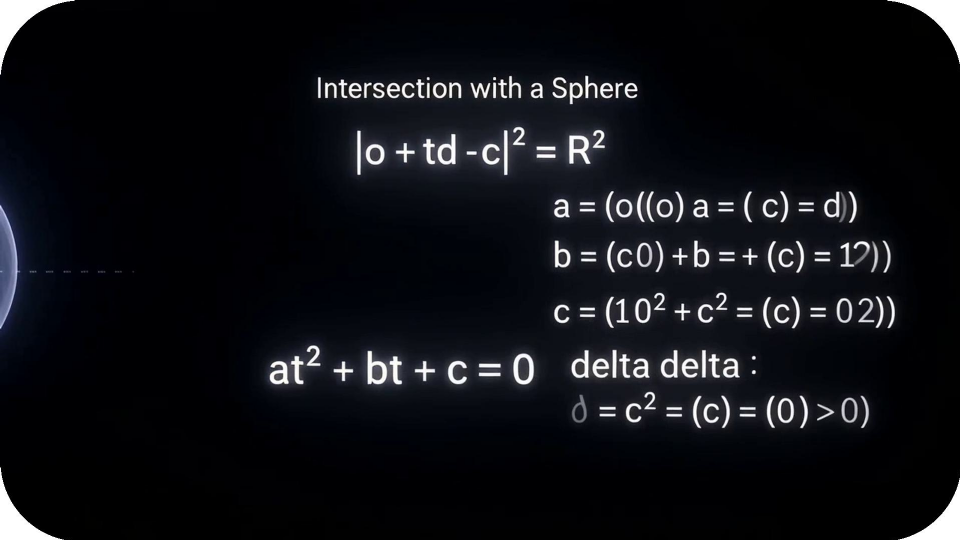}
        \caption{Pixel-level.}
        \label{fig:sora_pixel}
    \end{subfigure}
    \hfill
    \begin{subfigure}[t]{0.48\linewidth}
        \centering
        \includegraphics[width=\linewidth]{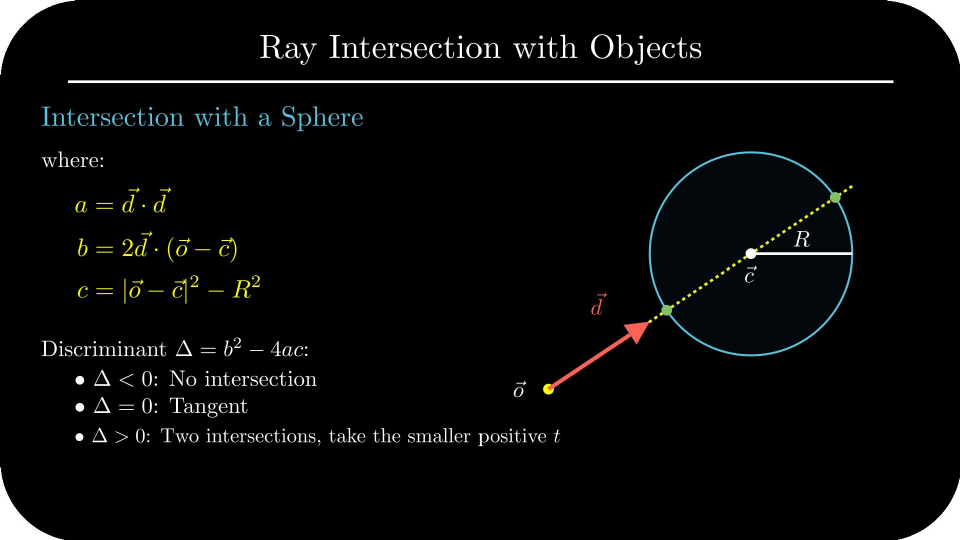}
        \caption{Programmatic.}
        \label{fig:sora_programmatic}
    \end{subfigure}

    \vspace{-0.5em}
    \caption{\textbf{Qualitative contrast between pixel-level and programmatic video generation.}}
    \label{fig:sora}
    \vspace{-1.1em}
\end{wrapfigure}

Recent advances in large language models (LLMs) and generative AI have broadened automated content creation from text and images to videos~\cite{li2025surveystateartlarge, yang2024cogvideoxtextvideodiffusionmodels, wu2025omnisvgunifiedgenerativeframework}. 
Automated video generation has since evolved along two major routes~\cite{yang2024cogvideoxtextvideodiffusionmodels}. 
Pixel-level methods, typically based on diffusion models, achieve impressive visual fidelity by modeling videos in pixel space~\cite{kong2024hunyuanvideosystematicframeworklarge, wang2025wanopenadvancedlargescale}. 
Without explicit geometric and symbolic constraints, they still struggle to maintain spatial consistency, accurate text and symbol rendering, and long-horizon logical coherence~\cite{wang2025motioncontrollablevideogenerationlatent, liu2025panowanliftingdiffusionvideo}. 
In contrast, programmatic methods execute scripts to define objects, layouts, transformations, and temporal dynamics~\cite{guan2025cadcodertextcadgenerationchainofthought, wei2025wordsstructuredvisualsbenchmark}. 
Grounded in executable code, this paradigm offers stronger geometric precision, controllability, and temporal consistency~\cite{chen2025code2video, xing2025empoweringllmsunderstandgenerate}. These strengths make it particularly suitable for knowledge-intensive scenarios such as educational animation, knowledge visualization, and scientific demonstration, where preserving spatial layouts, temporal dynamics, and symbolic semantics is more critical than merely achieving photorealistic appearance~\cite{wu2025omnisvgunifiedgenerativeframework, yang2025omnisvg, ku2025theoremexplainagentvideobasedmultimodalexplanations, chen-etal-2025-visualedu}, as demonstrated by the qualitative contrast between pixel-level and programmatic generation in Figure~\ref{fig:sora}.

Scaling programmatic video generation depends on whether current LLMs can synthesize video programs that are both executable and spatially and temporally correct~\cite{chen2025knowledgeenhancedlargelanguagemodels, qiu2025largelanguagemodelsunderstand, luo2026geogrambench, yao2025scipgnewbenchmarkapproach}. Existing benchmarks remain insufficient for systematically evaluating this capability~\cite{si2025design2codebenchmarkingmultimodalcode, wu2024plot2codecomprehensivebenchmarkevaluating}. First, existing visual code generation tasks mostly focus on static outputs, such as charts, webpages, SVGs, or single-frame layouts~\cite{yang2025chartmimicevaluatinglmmscrossmodal, zhao2025chartcoderadvancingmultimodallarge, koh2025c2scalableautofeedbackllmbased}, whereas programmatic video generation requires models to maintain consistency in object positions, relative relations, and motion trajectories over continuous time~\cite{wei2025wordsstructuredvisualsbenchmark, belouadi2025tikzerozeroshottextguidedgraphics}. Second, existing programmatic video datasets are typically small in scale~\cite{ku2025theoremexplainagentvideobasedmultimodalexplanations, oli2026trainingagenticinferencestrategies}, making it difficult to cover the diverse object compositions, spatial relations, and dynamic processes found in real-world scenarios. More importantly, existing evaluation protocols leave critical blind spots: code executability mainly measures whether a program can run~\cite{dong2025codescoreevaluatingcodegeneration, dong2025surveycodegenerationllmbased, zhao2025corebenchmarkingllmscodereasoning, qiao2025rethinkingverificationllmcodegeneration}, while coarse-grained VLM-as-judge evaluation provides only holistic judgments of the rendered video~\cite{pi2025mrjudgemultimodalreasoner, zhou2025projudgemultimodalmultidisciplinebenchmark, mondal2025countlooptrainingfreehighinstanceimage}. Neither can reliably capture fine-grained rendering failures that appear only in a few key frames, such as layout misalignment and out-of-bounds objects.

\vspace{-0.5em}
\begin{figure}[htbp]
  \centering
  \includegraphics[width=1\linewidth]{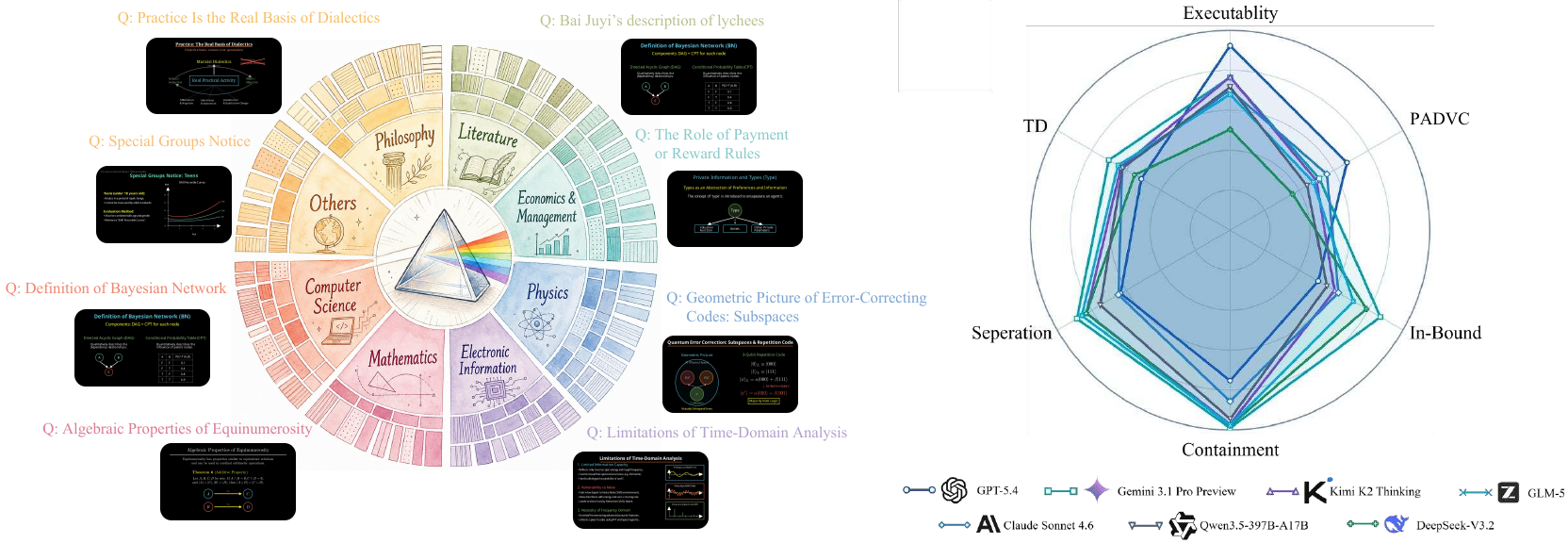}
  \caption{\textbf{Data overview of PRISM and aggregate model capability on the benchmark}. The left panel illustrates multi-level subject coverage with representative examples, while the right panel presents direction-aligned scores summarizing model capability.}
  \label{fig:overview}
\end{figure}
\vspace{-0.8em}

To address these issues, we introduce \textbf{PRISM} (\textbf{P}rogrammatic \textbf{R}easoning \textbf{I}n \textbf{S}patial \textbf{M}odalities) (See Figure~\ref{fig:overview}), a large-scale bilingual benchmark. PRISM instantiates programmatic video generation with Manim\footnote{Manim is a widely used Python package for programmatic video generation~\cite{manim2025code}.} and aims to characterize and bridge the \textit{Execution-Spatial Gap}: code may be executable, its rendered output can still contain spatial-layout and dynamic-process errors. Grounded in real scenarios, PRISM uses a hierarchical data curation framework that integrates task collection, code generation, automated filtering, and human revision. From more than 30,000 raw candidates, the pipeline filters out samples with execution failures, invalid spatial layouts, incomplete animations, or instruction-code mismatches, resulting in 10,372 executable, visually coherent, and instruction-aligned instruction-code pairs across English and Chinese subsets. These curated samples support reliable evaluation and provide high-quality data resources for future LLM training in programmatic video generation.

\vspace{-0.5em}

Our principal contributions are summarized as follows:

\begin{itemize}[leftmargin=*]
\vspace{-0.5em}
\item \textbf{Large-Scale Human-Calibrated Dataset}: We release PRISM, comprising 10,372 high-quality instruction-code pairs across English and Chinese subsets. Grounded in real scenarios and curated through a hierarchical quality-control framework, PRISM is $20\times$ larger than prior programmatic video generation benchmarks~\cite{ku2025theoremexplainagentvideobasedmultimodalexplanations, manimbench2026}.
\item \textbf{Multi-Dimensional Evaluation}: We design an automated evaluation framework with four complementary dimensions: \emph{Code-Level Reliability} measures execution robustness; \emph{Spatial Reasoning} assesses layout planning; \emph{Prompt-Aware Dynamic Visual Complexity} (PADVC) quantifies input-relevant dynamic richness; and \emph{Temporal Density} (TD) captures visual information density.
\item \textbf{Comprehensive LLM Benchmarking}: We conduct a large-scale and comprehensive evaluation of seven representative open- and closed-source LLMs across both subsets, complemented by further analyses and ablation studies. The results reveal a severe \textit{Execution-Spatial Gap}. Further analyses show that excessive dynamic intensity degrades spatial quality, and that extended thinking modes fail to bring reliable spatial gains, suggesting spatial planning as the key bottleneck.
\end{itemize}

\section{PRISM}
\label{sec:prism}

\subsection{Hierarchical Data Curation Framework}

PRISM is grounded in authentic knowledge visualization scenarios~\cite{chen-etal-2025-visualedu, chen2025code2video}. 
We collect English and Chinese outline fragments spanning diverse knowledge concepts and instantiate them as Manim programs. 
Manim is well suited to this setting, as it precisely maps Python instructions onto an absolute coordinate system, enabling explicit control over object placement, geometric relations, and temporal animations~\cite{guan2025cadcodertextcadgenerationchainofthought}. 
Although PRISM adopts Manim as its instantiation medium, its collection pipeline and subsequent evaluation protocol are decoupled from Manim, making the framework broadly generalizable~\cite{wei2025wordsstructuredvisualsbenchmark}. 
We then develop a collaborative data construction workflow that combines agent-based draft generation, annotator revision, and expert quality prescreening~\cite{ku2025theoremexplainagentvideobasedmultimodalexplanations, oli2026trainingagenticinferencestrategies}. 
This standardized production paradigm yields a raw pool of more than 30,000 candidate samples~\cite{chen2024spatialvlmendowingvisionlanguagemodels, li2024mvbenchcomprehensivemultimodalvideounderstanding}.

Since PRISM targets precise spatial relationships among objects across consecutive frames, including subtle failures such as unexpected occlusion and local out-of-frame placement, it requires a highly reliable, scalable quality-assurance mechanism.~\cite{fu2025videommefirstevercomprehensiveevaluationbenchmark, wang2024muirbenchcomprehensivebenchmarkrobust, zhang2025stvlmskinematicinstructiontuning}. VLMs are ill-suited to this role, given their known weaknesses in fine-grained spatiotemporal reasoning~\cite{tu2025odeopensetevaluationhallucinations, augustin2025dashdetectionassessmentsystematic}, susceptibility to hallucinations~\cite{li2025surveystateartlarge, sclar2025flawartifactrethinkingprompt}, and sensitivity to prompt variations~\cite{jia2025brittlebenchquantifyingllmrobustness, zhou2025robustnessreliabilitybenchmarkbasedevaluation}. We therefore construct a human-in-the-loop semi-automatic framework based on low-level rendering-engine parsing, relying on deterministic rules for core verification and manual review for boundary cases~\cite{wang2025spatial457, ogezi-shi-2025-spare, liu-etal-2025-multimodal-large, mayer-etal-2025-ivispar, zhang2025mitigatingspatialhallucinationlarge}.

\textbf{Step 1: Deterministic hard filtering based on low-level parsing.}
We instrument Manim's rendering pipeline directly. Automated scripts detect and quantify spatial layout violations based on exact coordinate data. Any candidate exhibiting geometric violations beyond defined thresholds in any frame is immediately removed. This stage eliminates the majority of samples with layout defects.

\textbf{Step 2: Boundary calibration with human priors.}
To avoid overly conservative rules that would unnecessarily reduce sample diversity, we construct a gold subset of 3,000 expert-annotated examples. We then analyze false positives and iteratively refine exemption rules and tolerance thresholds. 

\textbf{Step 3: Human spot-check final review.}
A small number of issues still require semantic inspection, such as whether the animation matches the educational content and whether the overall presentation is pedagogically appropriate.

This pipeline yields 10,372 high-quality instruction-code pairs in English and Chinese, each consisting of an educational outline segment and an executable Manim script.

\subsection{Task Definition}

We formulate PRISM as a constrained code generation task mapping natural language instructions to programmatic video. Formally:
\begin{equation}
f_{\theta}: I \rightarrow O, \quad I=\langle S, X \rangle, \quad O=(C, V)
\end{equation}
where $S$ specifies the rendering environment (e.g., version constraints) and $X$ provides educational context and semantic guidance within a broader teaching outline.

The output $O$ comprises a Manim program $C$ and its rendered video $V$. If execution fails, $V=\bot$; otherwise, $V$ is decomposed into a sequence of atomic animation events $\{e_1, \dots, e_k\}$, defined as indivisible rendering primitives, for evaluation. This formulation evaluates the model's ability to translate logical instructions into spatially consistent, long-sequence animations.

\subsection{Benchmark Statistics}

PRISM contains 10,372 instruction-code pairs, including 5,199 English and 5,173 Chinese. In terms of knowledge coverage, the benchmark spans 437 finer-grained subject categories.

\begin{figure}[htbp]
    \centering
    \hspace{-1.3cm}
    \includegraphics[width=1\linewidth]{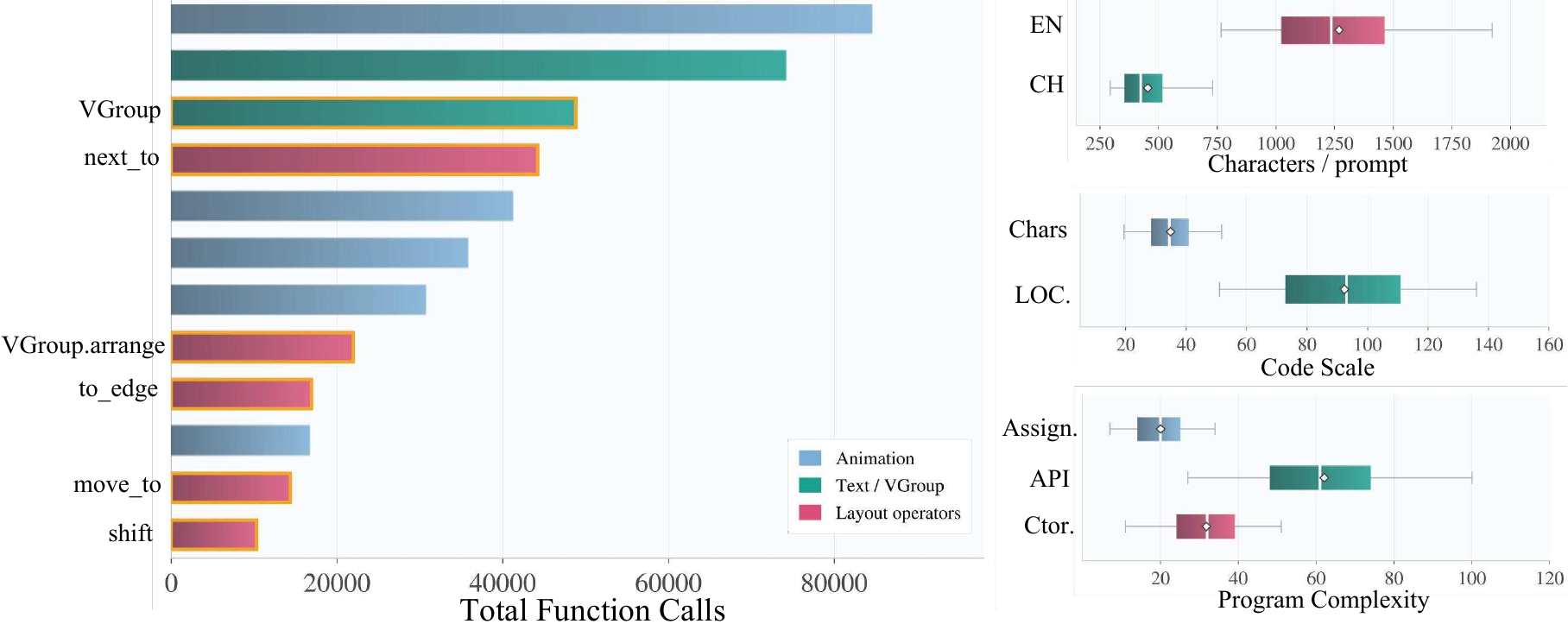}
    \caption{\textbf{Benchmark statistics}. Left: The 12 most frequently occurring Manim APIs and operators in the dataset. Top-right: Distribution of character counts per prompt. Middle-right: Scale of the reference code. Bottom-right: Composition of program structure.}
    \label{fig:benchmark_statistics}
\end{figure}

Inputs feature structured educational text averaging 1,169 characters (English) and 432 characters (Chinese). Over 87\% of samples include headings, lists, or LaTeX formulas, explicitly testing the model's ability to process complex content beyond simple prompts. Programmatically, reference scripts average 93 lines and 3,500 characters. AST analysis reveals highly event-dense logic: samples average 32 object constructions, 62 API calls, and 20 variable assignments. Although many scripts are organized around a single \texttt{Scene} class, their internal logic is highly event-dense.

Crucially, PRISM stresses spatial planning. Precise layout operators (e.g., \texttt{VGroup.arrange}, \texttt{next\_to}) significantly outnumber basic display primitives (\texttt{Text}, \texttt{MathTex}), testing whether models can construct geometrically coherent scenes rather than merely rendering isolated objects~\cite{guan2025cadcodertextcadgenerationchainofthought, wei2025wordsstructuredvisualsbenchmark}.

\section{Evaluation Methodology}
\label{sec:eval}

\begin{wrapfigure}{r}{0.36\linewidth}
    \vspace{-3.2em}
    \centering
    \includegraphics[width=\linewidth]{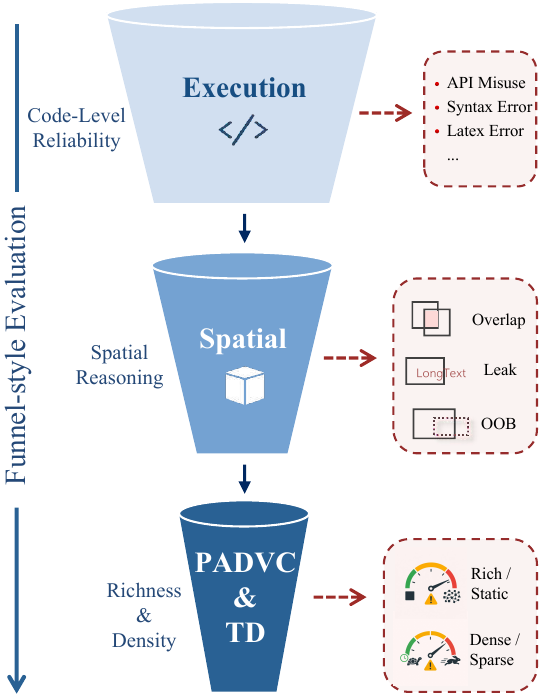}
    \caption{\textbf{Funnel-style evaluation framework}. From Code-Level Reliability to Spatial Reasoning, and to PADVC/TD diagnostic dimensions.}
    \label{fig:eval}
    \vspace{-4.4em}
\end{wrapfigure}

We construct a fine-grained evaluation suite spanning both code and visual dimensions, organized around four complementary metrics. \emph{Code-Level Reliability} measures execution robustness. \emph{Spatial Reasoning} assesses layout planning on a constrained two-dimensional canvas. \emph{Prompt-Aware Dynamic Visual Complexity} (PADVC) quantifies input-relevant dynamic richness, and \emph{Temporal Density} (TD) captures visual information density over time.

Together, these metrics form a funnel-style evaluation framework (Figure~\ref{fig:eval}). Code-Level Reliability acts as the outer gate: only samples that execute successfully enter downstream visual analysis, while failures are routed to fine-grained error diagnosis. Spatial Reasoning then serves as the core end-to-end metric, testing whether the model can maintain spatial coherence over the full rendered sequence. Finally, PADVC and TD function as diagnostic dimensions rather than primary ranking targets, characterizing how models differ in dynamic expression strength and temporal activity.

\subsection{Code-Level Reliability}

Code-level checking is the first step of the pipeline, since executable code is the necessary prerequisite for any downstream spatiotemporal and visual evaluation in a code-generation benchmark: any sample that fails execution is excluded from downstream spatiotemporal and visual evaluation. Note that samples which execute successfully but produce degenerate output (e.g., empty scenes) are retained and naturally penalized by the downstream metrics. Our code evaluation has three parts:

\textbf{Basic pass rate.} We use the execution success rate as the core indicator of basic code-generation capability, defined as the fraction of prompts for which the model's first attempt yields a video that the renderer completes without error.

\textbf{Fine-grained error categorization.} To diagnose failures more precisely, given the breadth of the rendering engine's API surface, we classify failures into six categories: API hallucination, API misuse, text-rendering error, formatting pollution, syntax error, and other. Each failed sample is mapped to its primary category.

\textbf{Execution complexity.} Beyond binary pass/fail, since runnable scripts may still differ substantially in computational cost, we also report the mean render time in the same controlled environment as a complementary indicator of computational cost, helping reveal whether a model tends to produce computationally expensive or redundant scripts.

\subsection{Spatial Reasoning}

Spatial reasoning evaluates a model's layout planning ability on a constrained two-dimensional canvas over long animation sequences. This aspect must be assessed explicitly because executable code does not necessarily preserve stable layout throughout the rendered sequence. We categorize spatial failures into three mutually exclusive modes:

\textbf{Layout overlap.} Multiple visual elements overlap or interfere in an unintended way. Intentional layering is excluded, we count occlusions that damage information delivery.

\textbf{Out-of-bounds placement.} The main body of a visual element extends beyond the default visible frame, causing physical information loss.

\textbf{Layout leakage.} The boundary or geometric extension of an element exceeds its proper local region, such as text escaping a text box or numbers protruding beyond matrix brackets.

For detection, to capture failures that accumulate across the sequence rather than only in a single final frame, we use an automated audit pipeline that captures key snapshots synchronized with animation actions, extracts each object's point set, bounding extent, and hierarchical relations via low-level interfaces, and deterministically checks all three spatial constraints. To avoid false alarms, the pipeline integrates two mechanisms: a hierarchy-aware primitive parser that recursively expands the scene graph and establishes parent-child relations, and a false-positive suppression module that uses topological and semantic cues to exempt intentional overlaps such as highlighting and grid adjacency. 

At the sample level, \emph{Spatial Pass} is counted only if a sample renders successfully and every audited stable frame over the full sequence passes all spatial checks, making the metric especially sensitive to layout degradation over long sequences.

\subsection{Prompt-Aware Dynamic Visual Complexity}

\begin{wrapfigure}{r}{0.52\linewidth}
  \vspace{-1em}
  \centering
  \begin{minipage}[t]{0.48\linewidth}
    \centering
    \includegraphics[width=\linewidth]{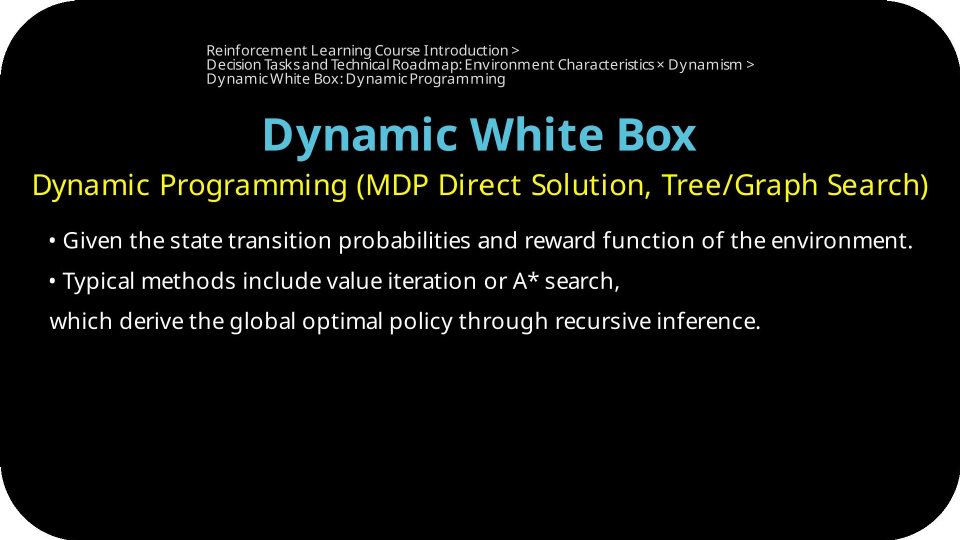}
    
    \vspace{0.25em}
    {\scriptsize \textbf{(a)} \textbf{Low PADVC.} Near-static and slideshow-like output.}
  \end{minipage}\hfill
  \begin{minipage}[t]{0.48\linewidth}
    \centering
    \includegraphics[width=\linewidth]{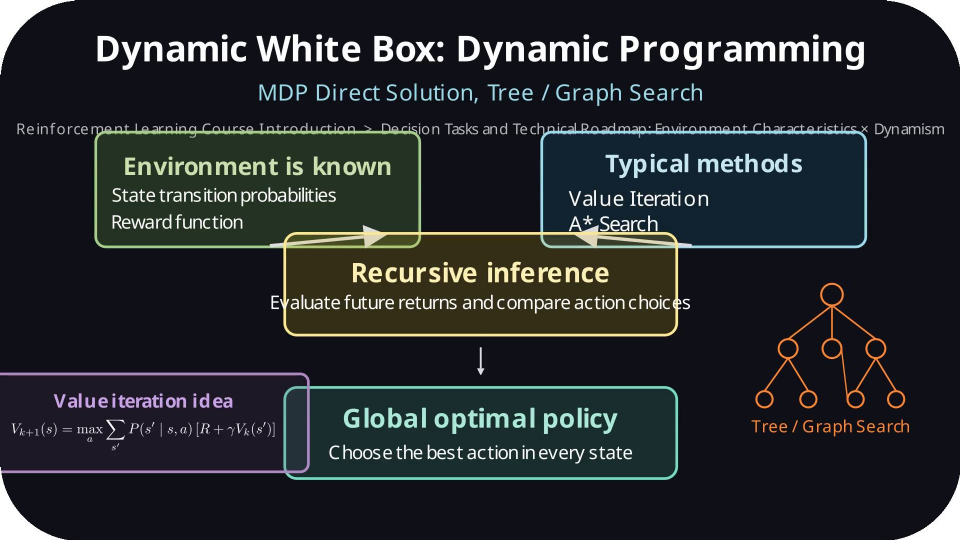}
    
    \vspace{0.25em}
    {\scriptsize \textbf{(b)} \textbf{High PADVC.} Excessive motion causes spatial instability.}
  \end{minipage}

  \vspace{-0.3em}
  \caption{\textbf{PADVC vs. generation quality.} Both under- and over-estimated dynamic visual complexity lead to failures.}
  \label{fig:case_study_padvc}
  \vspace{-2.1em}
\end{wrapfigure}

In practice, we observed two typical failure modes: models that produce near-static slideshow-like videos lacking dynamic process, and models that introduce excessive decorative motion likely to trigger spatial failures. To diagnose both, we introduce \emph{Prompt-Aware Dynamic Visual Complexity} (PADVC), which quantifies input-relevant dynamic visual complexity (see Figure~\ref{fig:case_study_padvc}).

PADVC rests on two principles: (1)~the model's visual contribution beyond reproducing input text is primarily reflected in the organization, evolution, and relational changes of non-text geometric objects rather than in text accumulation; (2)~a progressive presentation unfolding through semantically coherent steps is dynamically richer than a one-shot display. We define the raw score as
\begin{equation}
\mathrm{PADVC}_{r}
=
\frac{\sum_k \left(E_{\mathrm{geo}}^{(k)}\right)^p}
{\log(\mathrm{PVD}+e)\cdot\left(1+\left(E_{\mathrm{text}}^{\max}\right)^p\right)}
\end{equation}
where $E_{\mathrm{geo}}^{(k)}$ is the non-text geometric evolution energy of event $k$, $E_{\mathrm{text}}^{\max}$ is the peak textual visual burden over the full video, $\mathrm{PVD}$ denotes \emph{Prompt Visual Density}, i.e., the structured complexity of the input educational outline, $e$ is Euler's number, $\epsilon$ is a small constant for numerical stability, and $p\in(0,1)$ is a concave exponent that naturally rewards progressive, multi-step presentation over one-shot display. In practice, we compute $\mathrm{PVD}=N_{\mathrm{struct}}+N_{\mathrm{action}}$,
where $N_{\mathrm{struct}}$ counts structural markers in the input outline (e.g., headings, lists) and $N_{\mathrm{action}}$ counts action-oriented semantic cues.

Both $E_{\mathrm{text}}$ and $E_{\mathrm{geo}}$ are computed from frame-level image analysis. For each frame $t$, OCR detects text regions and produces a binary mask $\widehat{M}_{\mathrm{text}}(t)$. The frame-level text-boundary energy is
\begin{equation}
E_{\mathrm{text}}(t)
=
\sum \left| \Delta\!\left(\widehat{M}_{\mathrm{text}}(t)\right) \right|
\end{equation}
and the full-video textual burden is its peak: $E_{\mathrm{text}}^{\max}=\max_t E_{\mathrm{text}}(t)$.

For geometric evolution, we segment the video into animation events by thresholding the forward frame-change ratio. For event $k$ with start frame $I_{s_k}$ and end frame $I_{e_k}$, to capture construction rather than removal, we retain only newly revealed pixels via $D_+^{(k)}=\max(I_{e_k}-I_{s_k},0)$, then extract structural change and mask out text:
\begin{equation}
E_{\mathrm{geo}}^{(k)}
=
\sum
\left(
\left|\Delta\!\left(D_+^{(k)}\right)\right|
\odot
\left(1-\widetilde{M}_{\mathrm{text}}(e_k)\right)
\right)
\end{equation}
where $\odot$ is element-wise multiplication. $E_{\mathrm{text}}$ captures static textual burden and $E_{\mathrm{geo}}$ captures non-text structural changes, \(\mathrm{PADVC}_{\mathrm{r}}\) thus measures geometric dynamic expression relative to both textual burden and prompt complexity.

Because \(\mathrm{PADVC}_{\mathrm{r}}\) is non-monotonic, where excessively low values indicate near-static output and excessively high values suggest redundant motion, we define a centered score
\begin{equation}
\mathrm{PADVC}_{\mathrm{c}}
=
\exp\!\left(
-\frac{1}{2}
\left(
\frac{\log\!\left(\mathrm{PADVC}_{\mathrm{r}}+\epsilon\right)-\mu_{\mathrm{ref}}}
{\sigma_{\mathrm{ref}}}
\right)^2
\right),
\end{equation}
where $\mu_{\mathrm{ref}}$ and $\sigma_{\mathrm{ref}}$ are fitted from the distribution of $\log(\mathrm{PADVC}_{\mathrm{r}}+\epsilon)$ over the benchmark's ground-truth videos. Used jointly, \(\mathrm{PADVC}_{\mathrm{r}}\) reveals whether a model is under- or over-producing dynamics, while \(\mathrm{PADVC}_{\mathrm{c}}\) quantifies its proximity to reference-level dynamics.

\subsection{Temporal Density}

The temporal dimension involves two aspects: \emph{causal order} and \emph{temporal activity}. In our code-driven animation setting, where scene scripts are executed sequentially, causal order is generally preserved once Code-Level Reliability checks are passed. In our early experiments, we did not observe standalone temporal-order errors under this condition. Temporal activity, by contrast, still requires a dedicated metric.

Unlike PADVC, which focuses on non-text geometric dynamics, \emph{Temporal Density} (TD) measures the rate of frame-level visual change regardless of source. Let $r_t$ denote the fraction of pixels whose grayscale difference between consecutive frames exceeds a threshold $\tau$. The raw temporal density is
\begin{equation}
\mathrm{TD}_{\mathrm{r}}
=
\mathrm{fps}\cdot \frac{1}{T-1}\sum_{t=2}^{T} r_t,
\end{equation}
where $T$ is the total number of frames and the $\mathrm{fps}$ factor normalizes the per-frame change ratio to a per-second rate.

As with PADVC, \(\mathrm{TD}_{\mathrm{r}}\) is not monotonic. Very low values correspond to globally static videos, whereas excessively high values may indicate overly frequent refresh, redundant transitions, or large-area motion weakly related to the teaching target. We therefore define a centered version,
\begin{equation}
\mathrm{TD}_{\mathrm{c}}
=
\exp\!\left(
-\frac{1}{2}
\left(
\frac{\log\!\left(\mathrm{TD}_{\mathrm{r}}+\epsilon\right)-\mu_{\mathrm{TD}}}
{\sigma_{\mathrm{TD}}}
\right)^2
\right),
\end{equation}
where $\epsilon$ is the same smoothing constant used in PADVC, and $\mu_{\mathrm{TD}}$ and $\sigma_{\mathrm{TD}}$ are fitted from the reference-answer distribution.

\section{Experiments}
\label{sec:experiments}

\subsection{Experimental Setup}

\textbf{Models.}
We evaluate seven representative advanced models. The closed-source group includes GPT-5.4~\cite{openai2025gpt45systemcard}, Gemini 3.1 Pro Preview~\cite{comanici2025gemini25pushingfrontier}, Kimi K2 Thinking~\cite{moonshot2025kimik2openagentic}, and Claude Sonnet 4.6~\cite{anthropic2025claudeopus4claude}. The open-source group includes Qwen3.5-397B-A17B~\cite{yang2025qwen3technicalreport}, GLM-5~\cite{zhipu2025glm45agenticreasoningcoding}, and DeepSeek-V3.2~\cite{liu2025deepseekv32pushingfrontier}.

\textbf{Settings.}
All models are evaluated on the same randomly sampled 20\% subset of the benchmark. Because English and Chinese differ in text density and visual layout characteristics, leading to divergent metric distributions, the two languages are evaluated independently. We set the temperature to 0.7. Thinking mode is enabled where available~\cite{anthropic2025claudeextendedthinking}. All code is rendered in \texttt{Manim CE v0.19.0}, a mainstream stable release of Manim Community Edition\footnote{Manim CE v0.19.0 is the experiment anchor. PRISM will track mainstream updates.}. We set the PADVC concave exponent $p=0.7$. The centered PADVC and TD scores use language-specific reference-fit parameters estimated from the reference-answer distribution. For PADVC, the fitted $(\mu,\sigma)$ values are $(-0.6663, 0.6547)$ for Chinese and $(-2.4470, 1.8098)$ for English, with smoothing constant $10^{-8}$. For TD, the fitted $(\mu,\sigma,\epsilon)$ values are $(-3.6128, 0.5952, 9.81\times10^{-5})$ for Chinese and $(-3.4075, 0.4680, 4.71\times10^{-3})$ for English, and both languages use a frame-difference threshold of $\tau=25$.

\vspace{-0.5em}

\subsection{Main Results}

\begin{table*}[htbp]
\vspace{-0.5em}
\centering
\caption{\textbf{Main benchmark results}. \textbf{Exec.} (execution success rate) and \textbf{Spatial} (spatial pass rate) are reported as proportions; \textbf{PADVC} and \textbf{TD} report the centered scores $\mathrm{PADVC}_{\mathrm{c}}$ and $\mathrm{TD}_{\mathrm{c}}$; \textbf{Time} is mean render time in minutes; \textbf{Overlap}/\textbf{Leak}/\textbf{OOB.} are the three constituent spatial-failure rates, reported as proportions. Best results are in \textbf{bold}, runners-up are \underline{underlined}.}
\label{tab:main_results}
\renewcommand\tabcolsep{5.3pt}
\renewcommand\arraystretch{1.18}
\newcommand{\MainResultsHeaderStrut}{\rule[-0.45em]{0pt}{1.85em}}
\resizebox{\linewidth}{!}{%
\begin{NiceTabular}{>{\centering\arraybackslash}m{1.75cm}!{\color{black}\vrule width 0.6pt}>{\raggedright\arraybackslash}m{3.55cm}!{\color{black}\vrule width 0.6pt}cccccccc}
\CodeBefore
  \rectanglecolor{CadetBlue!20}{1-1}{2-10}
  \rowcolor{gray!15}{3,11}
  \rowcolor{gray!10}{5,7,9,13,15,17}
  \rectanglecolor{white}{4-1}{7-1}
  \rectanglecolor{white}{8-1}{10-1}
  \rectanglecolor{white}{12-1}{15-1}
  \rectanglecolor{white}{16-1}{18-1}
\Body
\arrayrulecolor{black}
\Xhline{0.6pt}
\multirowcell{2.5}[0pt][c]{\textbf{Source}} & \multirowcell{2.5}[0pt][c]{\textbf{Model}} & \multicolumn{4}{c!{\color{black}\vrule width 0.6pt}}{\MainResultsHeaderStrut\textbf{Generation Quality}} & \multicolumn{4}{c}{\MainResultsHeaderStrut\textbf{Visual Robustness}} \\
\Xcline{3-10}{0.6pt}
& & \multicolumn{1}{c!{\color{black}\vrule width 0.6pt}}{\MainResultsHeaderStrut\textbf{Exec.} $\uparrow$} & \multicolumn{1}{c!{\color{black}\vrule width 0.6pt}}{\MainResultsHeaderStrut\textbf{Time} $\downarrow$} & \multicolumn{1}{c!{\color{black}\vrule width 0.6pt}}{\MainResultsHeaderStrut\textbf{PADVC} $\uparrow$} & \multicolumn{1}{c!{\color{black}\vrule width 0.6pt}}{\MainResultsHeaderStrut\textbf{TD} $\uparrow$} & \multicolumn{1}{c!{\color{black}\vrule width 0.6pt}}{\MainResultsHeaderStrut\textbf{Spatial} $\uparrow$} & \multicolumn{1}{c!{\color{black}\vrule width 0.6pt}}{\MainResultsHeaderStrut\textbf{Overlap} $\downarrow$} & \multicolumn{1}{c!{\color{black}\vrule width 0.6pt}}{\MainResultsHeaderStrut\textbf{Leak} $\downarrow$} & \multicolumn{1}{c}{\MainResultsHeaderStrut\textbf{OOB.} $\downarrow$} \\
\Xhline{0.6pt}
\multicolumn{10}{c}{{\rule{0pt}{1.15em}\bfseries\itshape ENGLISH}} \\
\hline
\multirowcell{4}[0pt][c]{\textsc{Closed}} & \makecell[l]{GPT-5.4} & \textbf{0.945} & \underline{0.300} & \textbf{0.856} & 0.417 & 0.308 & 0.257 & 0.204 & 0.518 \\
& \makecell[l]{Gemini 3.1 Pro Preview} & 0.778 & 0.457 & 0.681 & \textbf{0.670} & \textbf{0.572} & \textbf{0.087} & \textbf{0.006} & \textbf{0.166} \\
& \makecell[l]{Kimi K2 Thinking} & 0.798 & 0.477 & \underline{0.690} & 0.619 & 0.280 & 0.143 & \underline{0.017} & 0.481 \\
& \makecell[l]{Claude Sonnet 4.6} & \underline{0.803} & 0.302 & 0.643 & 0.590 & 0.263 & 0.335 & 0.144 & 0.451 \\
\hline
\multirowcell{3}[0pt][c]{\textsc{Open}} & \makecell[l]{Qwen3.5-397B-A17B} & 0.530 & \textbf{0.245} & 0.537 & 0.473 & 0.258 & 0.154 & 0.020 & \underline{0.273} \\
& \makecell[l]{GLM-5} & 0.720 & 0.465 & 0.668 & \underline{0.645} & \underline{0.341} & \underline{0.127} & 0.019 & 0.349 \\
& \makecell[l]{DeepSeek-v3.2} & 0.796 & 0.512 & 0.640 & 0.572 & 0.250 & 0.224 & 0.058 & 0.523 \\
\hline
\multicolumn{10}{c}{{\rule{0pt}{1.15em}\bfseries\itshape CHINESE}} \\
\hline
\multirowcell{4}[0pt][c]{\textsc{Closed}} & \makecell[l]{GPT-5.4} & \textbf{0.902} & 0.202 & \textbf{0.496} & 0.610 & 0.177 & 0.478 & 0.289 & 0.463 \\
& \makecell[l]{Gemini 3.1 Pro Preview} & \underline{0.752} & \underline{0.152} & 0.351 & \textbf{0.733} & \textbf{0.557} & \textbf{0.141} & \textbf{0.013} & \textbf{0.095} \\
& \makecell[l]{Kimi K2 Thinking} & 0.737 & 0.298 & 0.307 & 0.643 & 0.331 & 0.195 & 0.039 & 0.311 \\
& \makecell[l]{Claude Sonnet 4.6} & 0.602 & \textbf{0.143} & \underline{0.477} & 0.658 & 0.171 & 0.373 & 0.142 & 0.296 \\
\hline
\multirowcell{3}[0pt][c]{\textsc{Open}} & \makecell[l]{Qwen3.5-397B-A17B} & 0.479 & 0.153 & 0.186 & 0.632 & 0.291 & 0.183 & \underline{0.022} & \underline{0.150} \\
& \makecell[l]{GLM-5} & 0.639 & 0.260 & 0.294 & 0.651 & \underline{0.340} & \underline{0.158} & 0.030 & 0.224 \\
& \makecell[l]{DeepSeek-v3.2} & 0.637 & 0.327 & 0.250 & \underline{0.672} & 0.204 & 0.275 & 0.052 & 0.356 \\
\Xhline{0.6pt}
\end{NiceTabular}}
\vspace{-1.0em}
\end{table*}

Among closed-source models, GPT-5.4 achieves the highest Exec. and PADVC in both languages yet suffers from high overlap and out-of-bounds rates, pointing to visually ambitious outputs that frequently violate spatial constraints~\cite{zhang2025mitigatingspatialhallucinationlarge}. Gemini 3.1 Pro Preview leads on Spatial, all three audit metrics, and TD, reflecting well-paced temporal activity within a spatially disciplined layout. Claude Sonnet 4.6 renders fastest but shows elevated overlap and leakage rates. Among open-source models, GLM-5 is the most balanced, outperforming Claude Sonnet 4.6 on Spatial in both languages. DeepSeek-V3.2 is more prone to boundary-control instability, whereas Qwen3.5-397B-A17B, despite its lower Exec., maintains comparatively low audit-issue rates once the code runs successfully. Overall, closed-source models tend to lead on code-level reliability, but this advantage does not uniformly extend to spatial robustness.

\textbf{Execution-Spatial Gap}. The most salient pattern is a persistent gap between Exec. and Spatial: even when code renders successfully, the proportion passing all spatial checks drops sharply, with a mean Exec.-to-Spatial decline of 41.24 percentage points (95\% bootstrap CI [40.41, 42.11]). This confirms that the core difficulty lies not in producing runnable Manim code but in maintaining coherent spatial organization over long animated sequences~\cite{quan2025stark}.

\begin{figure}[htbp]
    \centering
    \includegraphics[width=0.9\linewidth]{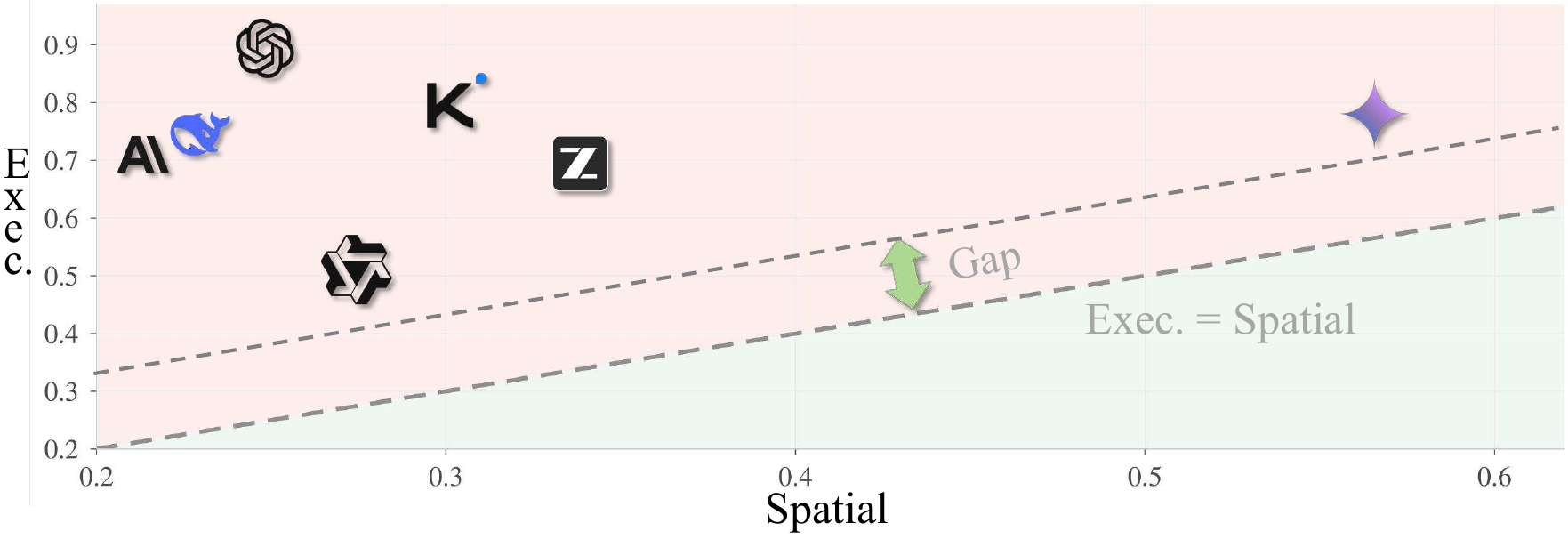}
    \caption{\textbf{Execution-Spatial Gap}. All models lie above the diagonal, indicating a clear disconnect between execution success and spatial pass.}
    \label{fig:padvc_td_visual}
    \vspace{-1.5em}
\end{figure}

The gap size varies revealingly across models. Gemini 3.1 Pro Preview pairs the strongest Spatial scores with one of the smallest gaps, suggesting that its code generation and spatial planning are well aligned. Qwen3.5-397B-A17B has the lowest Exec. yet a small gap, pointing to stable spatial control once code compiles. GPT-5.4 presents the opposite profile: the highest Exec. yet the largest gap in both languages, confirming that code-generation strength does not transfer to spatial robustness~\cite{zhang2025mitigatingspatialhallucinationlarge}.

\vspace{-0.3em}

\subsection{Further Analysis}

\textbf{Code-side failure patterns}. Table~\ref{tab:code_error_breakdown_core} shows that code-level failures are dominated not by basic syntax errors but by Manim-specific knowledge gaps and rendering-backend instability. Syntax errors and formatting pollution account for only a small part of the failure mass for most models, so the most discriminative categories are API hallucination, API misuse, and text-rendering errors. API misuse is particularly prominent among open-source models: Qwen3.5-397B-A17B reaches 34.7 points in English and 33.9 points in Chinese, with GLM-5 and DeepSeek-V3.2 showing weaker but similar tendencies. Gemini 3.1 Pro Preview is more prone to API hallucination than GPT-5.4, while text-rendering errors rise sharply in Chinese for Kimi K2 Thinking and GLM-5.

\vspace{-1em}

\begin{table}[htbp]
\centering
\caption{\textbf{Compact code-side error breakdown}. Exec. Fail reports the percentage of samples that fail to execute, while the remaining columns show the failure components most relevant.}
\label{tab:code_error_breakdown_core}
\footnotesize
\renewcommand\tabcolsep{3.8pt}
\renewcommand\arraystretch{1.12}
\newcommand{\CodeErrCoreHeaderStrut}{\rule[-0.42em]{0pt}{1.68em}}
\resizebox{\linewidth}{!}{%
\begin{NiceTabular}{>{\centering\arraybackslash}m{1.55cm}!{\color{black}\vrule width 0.6pt}>{\raggedright\arraybackslash}m{3.50cm}!{\color{black}\vrule width 0.6pt}>{\centering\arraybackslash}m{1.6cm}!{\color{black}\vrule width 0.6pt}>{\centering\arraybackslash}m{1.15cm}>{\centering\arraybackslash}m{1.55cm}>{\centering\arraybackslash}m{1.65cm}}
\CodeBefore
  \rectanglecolor{CadetBlue!20}{1-1}{1-6}
  \rowcolor{gray!15}{2,10}
  \rowcolor{gray!10}{4,6,8,12,14,16}
  \rectanglecolor{white}{3-1}{6-1}
  \rectanglecolor{white}{7-1}{9-1}
  \rectanglecolor{white}{11-1}{14-1}
  \rectanglecolor{white}{15-1}{17-1}
\Body
\arrayrulecolor{black}
\Xhline{0.6pt}
\CodeErrCoreHeaderStrut\textbf{Source} 
& \multicolumn{1}{c!{\color{black}\vrule width 0.6pt}}{\CodeErrCoreHeaderStrut\textbf{Model}}
& \CodeErrCoreHeaderStrut\textbf{Exec. Fail}
& \CodeErrCoreHeaderStrut\textbf{Halluc.}
& \CodeErrCoreHeaderStrut\textbf{API Misuse}
& \CodeErrCoreHeaderStrut\textbf{Text Render} \\
\Xhline{0.6pt}
\multicolumn{6}{c}{{\rule{0pt}{1.15em}\bfseries\itshape ENGLISH}} \\
\hline
\multirowcell{4}[0pt][c]{\textsc{Closed}} & \makecell[l]{GPT-5.4} & \textbf{5.5} & \textbf{0.3} & \textbf{1.6} & \textbf{1.1} \\
& \makecell[l]{Gemini 3.1 Pro Preview} & 22.2 & 7.2 & 6.5 & 5.6 \\
& \makecell[l]{Kimi K2 Thinking} & 20.3 & \underline{1.2} & 5.7 & 2.6 \\
& \makecell[l]{Claude Sonnet 4.6} & \underline{19.8} & 4.5 & \underline{3.7} & \underline{1.2} \\
\hline
\multirowcell{3}[0pt][c]{\textsc{Open}} & \makecell[l]{Qwen3.5-397B-A17B} & 47.0 & 6.6 & 34.7 & 2.3 \\
& \makecell[l]{GLM-5} & 28.1 & 6.7 & 15.1 & 3.0 \\
& \makecell[l]{DeepSeek-V3.2} & 20.3 & 1.3 & 4.7 & 3.8 \\
\hline
\multicolumn{6}{c}{{\rule{0pt}{1.15em}\bfseries\itshape CHINESE}} \\
\hline
\multirowcell{4}[0pt][c]{\textsc{Closed}} & \makecell[l]{GPT-5.4} & \textbf{9.8} & \textbf{0.6} & \textbf{1.0} & \textbf{2.6}\\
& \makecell[l]{Gemini 3.1 Pro Preview} & \underline{24.8} & 6.4 & 8.0 & \underline{4.7} \\
& \makecell[l]{Kimi K2 Thinking} & 26.4 & \underline{2.0} & 5.5 & 14.1 \\
& \makecell[l]{Claude Sonnet 4.6} & 39.9 & 2.5 & \underline{4.5} & 6.3 \\
\hline
\multirowcell{3}[0pt][c]{\textsc{Open}} & \makecell[l]{Qwen3.5-397B-A17B} & 52.1 & 7.2 & 33.9 & 6.1 \\
& \makecell[l]{GLM-5} & 36.2 & 4.8 & 10.2 & 14.1 \\
& \makecell[l]{DeepSeek-V3.2} & 36.3 & 2.8 & 12.6 & 12.6 \\
\Xhline{0.6pt}
\end{NiceTabular}}

\end{table}

\textbf{Relationship between PADVC, TD, and Spatial}. $\mathrm{PADVC}_{\mathrm{c}}$ and $\mathrm{TD}_{\mathrm{c}}$ measure style alignment to the reference distribution, while Spatial evaluates whether dynamics fit in a stable layout. These metrics can diverge: GPT-5.4 leads on English $\mathrm{PADVC}_{\mathrm{c}}$ but not Spatial, whereas Gemini achieves the best Spatial without the highest dynamic scores. High $\mathrm{PADVC}_{\mathrm{r}}$ signals overload, with top-decile samples passing Spatial at 32\% versus 60\% for the bottom decile. TD reverses this trend: the lowest $\mathrm{TD}_{\mathrm{c}}$ decile reaches only 32\% Spatial, while the highest reaches 48\%. Excessive dynamics hurt Spatial, while calibrated pacing helps. Figure~\ref{fig:case_study_padvc} shows this contrast: low-$\mathrm{PADVC}_{\mathrm{r}}$ outputs become near-static slideshows, while high-$\mathrm{PADVC}_{\mathrm{r}}$ ones exhaust the spatial budget and trigger layout violations.

\textbf{Textual expansion and spatial quality}. Complementing PADVC, we define a lightweight diagnostic, TextExpand, to measure a model's tendency to expand the input into additional on-screen text:
\vspace{-0.2em}
\[
\mathrm{TextExpand}
=
\frac{T_{\mathrm{display}}^{\mathrm{uniq}}}
{T_{\mathrm{prompt}}}
\]

\vspace{-0.7em}where $T_{\mathrm{display}}^{\mathrm{uniq}}$ is the number of unique on-screen text tokens extracted from the generated code via syntax-aware parsing, and $T_{\mathrm{prompt}}$ is the token count of the input prompt. Overall, the linear correlation between $\mathrm{TextExpand}$ and Spatial is weak (Pearson $r=-0.048$), indicating that moderate expansion does not systematically damage spatial quality. The effect concentrates in the tail: the top $\mathrm{TextExpand}$ quintile achieves a Spatial rate of only $36.6\%$, compared with $49\%$--$54\%$ for the middle ranges. This tail effect is sharper in Chinese, where the highest quintile drops to $32.1\%$ while the second quintile reaches $59.5\%$. Excessive text expansion thus acts as a risk factor for spatial failure, particularly in Chinese, but moderate levels are benign~\cite{zhang2025mitigatingspatialhallucinationlarge}.
\begin{figure}[htbp]
  \centering
  \begin{minipage}[h]{0.49\linewidth}
    \centering
    \includegraphics[width=\linewidth]{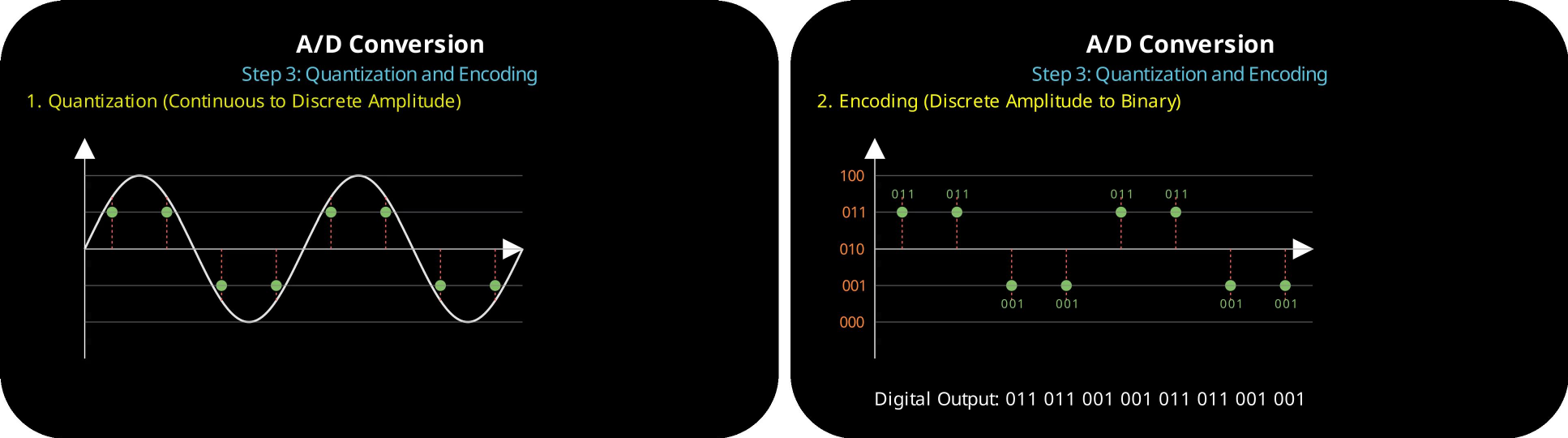}
    
    \vspace{0.35em}
    {\small\textbf{(a) Gemini 3.1 Pro Preview}}
  \end{minipage}\hfill
  \begin{minipage}[h]{0.49\linewidth}
    \centering
    \includegraphics[width=\linewidth]{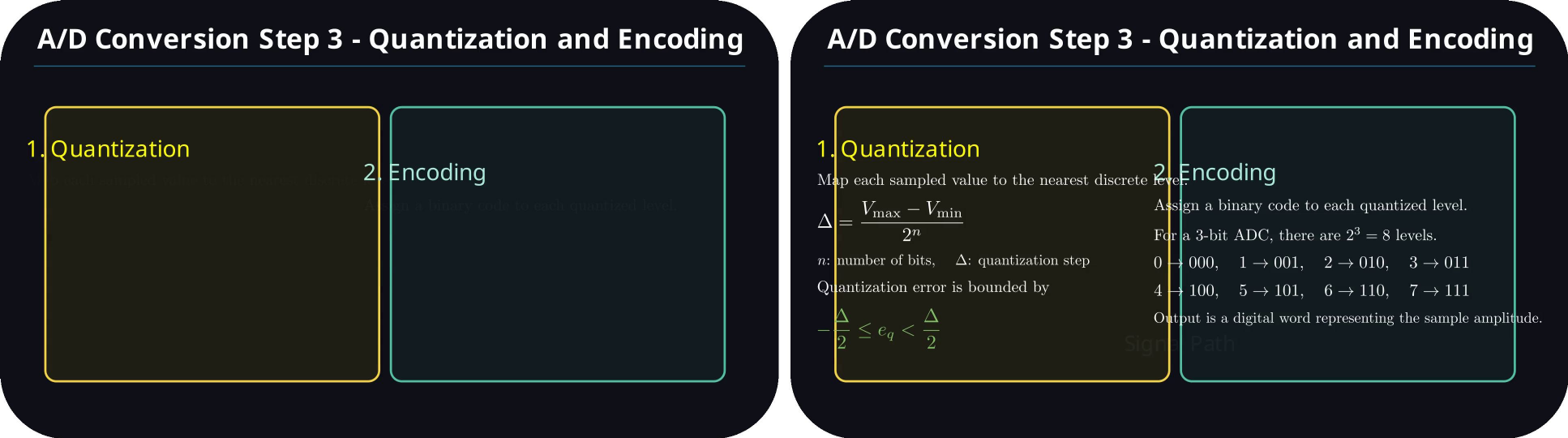}
    
    \vspace{0.35em}
    {\small\textbf{(b) GPT-5.4}}
  \end{minipage}
  \caption{\textbf{TextExpand cases}. The left Gemini sample remains concise, whereas the right GPT-5.4 sample expands visible text aggressively and fails under a substantially heavier spatial burden.}
  \label{fig:case_study_textexpand}
  \vspace{-1.5em}
\end{figure}

\textbf{Joint diagnosis with PADVC and total energy}. We further compare a collapsed total-energy score against the separated view of PADVC and TextExpand. Total energy does identify high-risk outputs, with the top 10\% reaching an 80.4\% Spatial failure rate overall. However, combining PADVC and TextExpand as separate axes yields a more precise high-risk region: samples that lie in both the top PADVC 10\% and the top TextExpand 10\% fail Spatial 83.0\% overall and 87.7\% in Chinese, compared with 82.5\% overall and 79.0\% in Chinese for a same-size total-energy selection. We therefore keep PADVC and TextExpand separated in the main analysis: the former diagnoses non-text dynamic overload, while the latter isolates textual expansion burden.

\begin{wrapfigure}{r}{0.5\linewidth}
    \vspace{-5pt} 
    \centering
    \includegraphics[width=\linewidth]{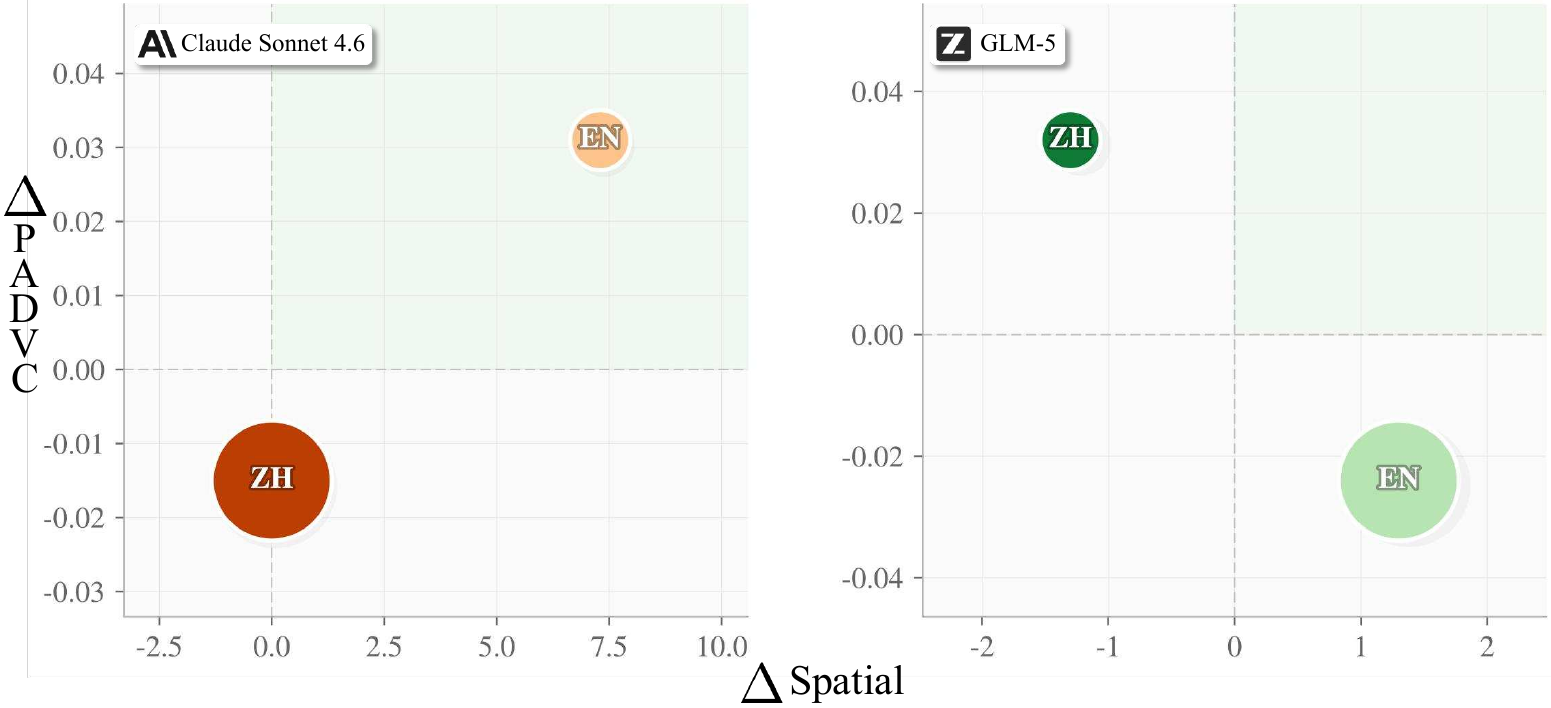}
\caption{\textbf{Thinking ablation across models and languages}. Deltas are computed as thinking minus base. $\Delta$Spatial (x) vs.\ $\Delta \text{PADVC}_c$ (y). Bubble size indicates latency increase; color intensity encodes token increase. The green quadrant marks simultaneous improvement.}
    \label{fig:thinking}
    \vspace{-15pt} 
\end{wrapfigure}

\begin{figure}[htbp]
\vspace{-0.5em}
  \centering
  \begin{minipage}[t]{0.49\linewidth}
    \centering
    \includegraphics[width=\linewidth]{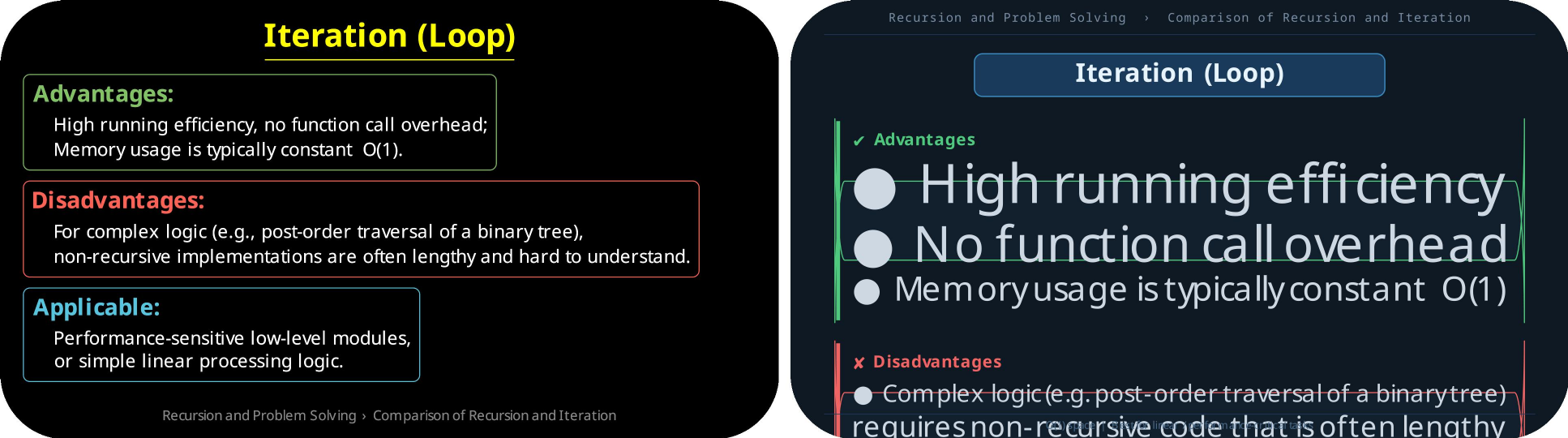}
    
    \vspace{0.35em}
    {\small\textbf{(a) Claude Sonnet 4.6}}
  \end{minipage}\hfill
  \begin{minipage}[t]{0.49\linewidth}
    \centering
    \includegraphics[width=\linewidth]{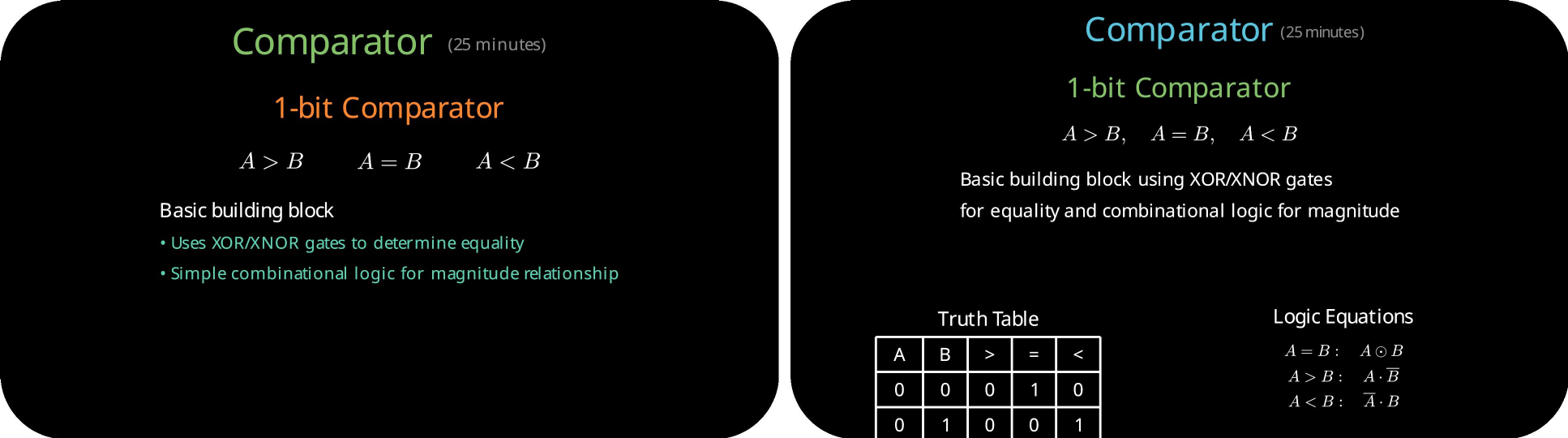}
    
    \vspace{0.35em}
    {\small\textbf{(b) GLM-5}}
  \end{minipage}
  \caption{\textbf{Thinking-ablation cases}. Left denotes the base and right denotes the thinking mode.}
  \label{fig:thinking_case_iteration}
  \vspace{-1.5em}
\end{figure}

\textbf{Effect of thinking}. Figure~\ref{fig:thinking} shows that enabling thinking mode brings inconsistent changes for Claude and GLM. For Claude, Spatial improves in English ($\pm7.3$\,\%) but is unchanged in Chinese, while Exec. changes in opposite directions across languages. For GLM, Spatial shifts by only $\pm1.3$\,\% despite five extra minutes per sample. Changes in $\mathrm{PADVC}_{\mathrm{c}}$ also do not track spatial gains, suggesting that extended reasoning does not reliably improve spatial organization~\cite{ogezi-shi-2025-spare}. Figure~\ref{fig:thinking_case_iteration} shows same-prompt examples from Claude Sonnet 4.6 and GLM-5. In both cases, thinking mode makes bolder layout changes: Claude enlarges text and containers, causing overlap and out-of-bounds errors, while GLM adds lower regions pushed out of frame. Thus, thinking may alter the reasoning strategy, but it does not ensure spatially disciplined animations.

\section{Related Work}

\vspace{-0.5em}

\textbf{Programmatic visual generation}. Pixel-level video generation with diffusion models achieves high visual fidelity but lacks the structural precision required for knowledge visualization, where geometric accuracy is critical~\cite{xing2025empoweringllmsunderstandgenerate}. Programmatic visual generation instead uses explicit code to ensure deterministic rendering and precise coordinate control~\cite{yang2025omnisvg}. Most benchmarks in this paradigm target static, single-frame outputs: Design2Code~\cite{si2025design2codebenchmarkingmultimodalcode} evaluates HTML/CSS generation, Plot2Code~\cite{wu2024plot2codecomprehensivebenchmarkevaluating} and ChartMimic~\cite{yang2025chartmimicevaluatinglmmscrossmodal} focus on chart reproduction, and TikZ benchmarks~\cite{wei2025wordsstructuredvisualsbenchmark} assess scientific figure generation. These works are limited to instantaneous layouts and overlook multi-step animation sequences, where spatial constraints must persist as scenes evolve~\cite{zhao2025chartcoderadvancingmultimodallarge, koh2025c2scalableautofeedbackllmbased, rahman-etal-2025-text2vis}. Within the Manim ecosystem, existing datasets are small, lack spatial quality control, and mainly evaluate execution success~\cite{belouadi2025tikzerozeroshottextguidedgraphics, qiu2025largelanguagemodelsunderstand, zhu2026autofigure, ku2025theoremexplainagentvideobasedmultimodalexplanations, manimbench2026, chen-etal-2025-visualedu, chen2025code2video}.

\textbf{Spatial Reasoning in Language Models}. Spatial reasoning has been identified as a persistent weakness of LLMs~\cite{yang2025thinkinginspace, zhang2025sphere, ogezi-shi-2025-spare}. Text-based probes~\cite{premsri2025forest, xu2025definingandevaluating} and visual benchmarks~\cite{quan2025stark, gupta2025evaluatingspatial} consistently show that models struggle with relative positioning and coordinate reasoning, but these evaluations operate in a \emph{perceptual} setting where the model observes an image or description and answers spatial queries~\cite{wang2025spatial457, liu-etal-2025-multimodal-large, mayer-etal-2025-ivispar}. PRISM tests a harder, \emph{generative} form: the model must produce code whose execution yields geometrically correct layouts over extended sequences, without observing the rendered output during generation~\cite{luo2026geogrambench, wang2025skelayout}.
\vspace{-0.8em}
\section{Conclusion}

\vspace{-0.5em}

We introduced PRISM, a large-scale benchmark for evaluating programmatic video generation under long-horizon spatial constraints. Built from 10,372 human-calibrated English-Chinese instruction-code pairs, PRISM combines realistic knowledge-visualization prompts with a deterministic renderer-grounded evaluation framework covering code reliability, spatial reasoning, dynamic visual complexity, and temporal activity. Experiments on seven representative LLMs reveal a pronounced Execution-Spatial Gap, with an average drop of about 41\% from execution success to spatial pass rate, showing that runnable code does not necessarily produce spatially coherent animations and that long-horizon spatial reasoning remains the key bottleneck. We hope PRISM will serve as a standardized testbed for future research on spatially grounded programmatic video generation.

\bibliographystyle{plainnat}
\bibliography{references}

\end{document}